\documentclass[10pt,twocolumn,letterpaper]{article}

\usepackage{wacv}
\usepackage{times}
\usepackage{epsfig}
\usepackage{graphicx}
\usepackage{amsmath}
\usepackage{amssymb}

\def\wacvPaperID{376} % Enter the WACV Paper ID here

\wacvfinalcopy % *** Uncomment this line for the final submission

\ifwacvfinal
\def\assignedStartPage{1} % *** Enter the assigned starting page number (instead of 9876)
\fi

\ifwacvfinal
\usepackage[breaklinks=true,bookmarks=false]{hyperref}
\else
\usepackage[pagebackref=true,breaklinks=true,colorlinks,bookmarks=false]{hyperref}
\fi

\ifwacvfinal
\else
\fi

\begin{document}

%%%%%%%%% TITLE
\title{ResNet or DenseNet? Introducing Dense Shortcuts to ResNet}

\author{Chaoning Zhang \\
% KAIST, South Korea\\
%5{\tt\small chaoningzhang1990@gmail.com}
\and
Philipp Benz \\
%KAIST, South Korea\\
%{\tt\small pbenz@kaist.ac.kr}
\and
Dawit Mureja Argaw \\
%KAIST, South Korea\\
%{\tt\small dawitmureja@kaist.ac.kr}
\and
Seokju Lee \\
%KAIST, South Korea\\
%{\tt\small seokju91@gmail.com}
\and
Junsik Kim \\
%KAIST, South Korea\\
%{\tt\small mibastro@gmail.com}
\and
Francois Rameau \\
%KAIST, South Korea\\
%{\tt\small rameau.fr@gmail.com}
\and

Jean-Charles Bazin \\
%KAIST, South Korea\\
%{\tt\small bazinjc@kaist.ac.kr}
\and
In So Kweon \\
%KAIST, South Korea\\
%{\tt\small iskweon@kaist.ac.kr}
\and
Korea Advanced Institute of Science and Technology \\
{\tt\small Contact email: chaoningzhang1990@gmail.com}
}

% \and

\maketitle
%\thispagestyle{empty}

%%%%%%%%% ABSTRACT
\begin{abstract}
ResNet or DenseNet? Nowadays, most deep learning based approaches are implemented with seminal backbone networks, among them the two arguably most famous ones are ResNet and DenseNet. Despite their competitive performance and overwhelming popularity, inherent drawbacks exist for both of them. For ResNet, the identity shortcut that stabilizes training also limits its representation capacity, while DenseNet has a higher capacity with multi-layer feature concatenation. However, the dense concatenation causes a new problem of requiring high GPU memory and more training time. Partially due to this, it is not a trivial choice between ResNet and DenseNet. This paper provides a unified perspective of dense summation to analyze them, which facilitates a better understanding of their core difference. We further propose dense weighted normalized shortcuts as a solution to the dilemma between them. Our proposed dense shortcut inherits the design philosophy of simple design in ResNet and DenseNet. On several benchmark datasets, the experimental results show that the proposed DSNet achieves significantly better results than ResNet, and achieves comparable performance as DenseNet but requiring fewer computation resources. 

\end{abstract}

\section{Introduction}
Deep neural networks (DNNs) have achieved state-of-the-art performance in numerous computer vision tasks~\cite{he2016deep,huang2017densely,girshick2015fast,ren2015faster,girshick2014rich,long2015fully,zhang2019revisiting,zhang2020deepptz,feng2020adversarial}, and the interpretation of DNNs has also be investigated from the lens of visualization~\cite{zeiler2014visualizing,mahendran2015understanding,olah2017feature} as well as robustness~\cite{szegedy2013intriguing,goodfellow2014explaining,benz2020data,zhang2019cd-uap,zhang2020understanding,benz2020double,bai2020targeted}. AlexNet~\cite{krizhevsky2012imagenet} and VGGNet~\cite{simonyan2014very} were the pioneering works that demonstrated the potential of DNNs. Inspired by the success of these seminal works, the research focus of the community has shifted from feature engineering~\cite{lowe2004distinctive,dalal2005histograms} to network design engineering and numerous new network architectures have come out to boost the performance of DNNs. ResNet, reusing preceding features with the identity shortcut had large success and achieved state-of-the-art performance on several benchmark datasets, such as ImageNet~\cite{deng2009imagenet} and COCO detection dataset~\cite{lin2014microsoft}. Compared to the Inception family of architectures, including GoogleNet~\cite{szegedy2015going} and Inception-V3~\cite{szegedy2016rethinking}, ResNet family shows better generalization, which implies that the learned features can be used in transfer learning with better efficiency~\cite{zagoruyko2016wide}. 

One of the reasons that make ResNet exceptionally popular is the simple design strategy which introduces only one identity shortcut. Despite its large success, the weakness of the identity shortcut has been analyzed by follow-up works. The identity shortcut skips the residual blocks to preserve features, as a result limiting the representation power of the network~\cite{zagoruyko2016wide,zhang2019revisiting}. The drawback of ResNets is that it causes the collapsing domain problem which reduces the learning capacity of the network~\cite{philipp2017gradients} and ~\cite{zhang2019revisiting} proposed to mitigate it with non-linear shortcuts. 

Another simple yet effective technique, dense concatenation, has been proposed in DenseNet~\cite{huang2017densely} to facilitate training deep networks. The DenseNet adopting dense concatenation to all subsequent layers to avoid using direct summation, preserves the features in preceding layers. DenseNet has been shown to have better feature use efficiency, outperforming ResNet with fewer parameters~\cite{huang2017densely}. Nonetheless, DenseNet requires heavy GPU memory due to concatenation operations. The memory issue can be mitigated by memory-efficient implementation introduced in~\cite{pleiss2017memory}. However, such an implementation is more complex from the engineering perspective, and it also further increases the training time of DenseNet by $20\%$~\cite{pleiss2017memory}. The main reason that DenseNet requires more training time is that DenseNet uses many small convolutions in the network, which runs slower on GPU than compact large convolution with the same number of GFLOPS. In short, there is a dilemma in the choice between ResNet and DenseNet for broad applications in terms of the performance and GPU resources. 

This paper proposes a dense normalized shortcut as an alternative dense connection technique to mitigate this dilemma. The proposed dense normalized shortcut outperforms ResNet~\cite{he2016deep} with a significant margin, with negligible parameter overhead, and it achieves comparable performance as DenseNet but requiring less computation. Our proposed network structure adopts the same backbone (convolutional block design) as ResNet and replaces identity shortcut with our dense normalized shortcut.
Our approach uses neither identity shortcut nor dense concatenation. From this perspective, this work is most similar to FractalNet~\cite{larsson2016fractalnet} whose structural layouts are precisely fractal. To our best knowledge, FractalNet is the only work that explored to train deep networks using neither identity shortcut nor dense concatenation, however, its performance is less favourable. The non-linear shortcuts introduced in~\cite{zhang2019revisiting} leads to performance boost. However, without idenity shortcut, its performance decreases when the network is very deep.

Overall, this paper provides one unified perspective of dense summation to analyze ResNet and DenseNet, which facilitates a better understanding of their core differences. Based on this perspective, we propose dense weighted normalized shortcuts to alleviate the drawbacks of the existing two dense connection techniques. We evaluate the proposed DSNet on several benchmark datasets and the results show that it outperforms ResNet by a significant margin and also with fewer parameters achieves comparable (or slightly better) performance as DenseNet but requiring fewer computation resources.

\section{Related works}
Deep CNN network design has become a very hot research topic and numerous techniques contributed to the success of deep learning in the computer vision field. Those techniques can be roughly divided into two categories: micro-module design and macro-architecture design.
\subsection{Micro-module design}
Micro-modules, such as normalization modules~\cite{ioffe2015batch}, attention modules~\cite{hu2018squeeze}, group convolutions~\cite{zhang2017interleaved}, and bottleneck design~\cite{he2016deep}, can be inserted into existing macro-architecture networks to improve the performance. Among them, normalization techniques~\cite{ioffe2015batch} are the most widely used and by default almost all the deep learning models adopt batch normalization~\cite{ioffe2015batch} to improve the performance and speed up the convergence. The random sampling in batch normalization also contributes to improving the generalization capability of the model. Recently, it has been shown in~\cite{benz2020batch} that batch normalization might increase adversarial vulnerability and updating the moving average statistics is found to improve the model corruption robustness. Alternative normalization techniques, such as weight normalization~\cite{salimans2016weight}, instance normalization~\cite{vedaldi2016instance} or layer normalization~\cite{ba2016layer}, have been investigated for addressing the dependency between samples during training. Those alternative techniques can also speed up the convergence, however, often do not provide as good performance as batch normalization. Both instance normalization and layer normalization can be seen as a special case of the later proposed group normalization~\cite{wu2018group}. Similar to instance normalization and layer normalization, group normalization performs the normalization along the channel direction instead of batch direction, enabling it to work effectively in memory-intensive applications where only a small number of samples can be processed in one batch~\cite{wu2018group}. In previous works, the normalization techniques have been mainly used in the residual path but our work explores the effect of the normalization techniques in the dense shortcut path. Normalization in the non-linear shortcut has also been investigated in~\cite{zhang2019revisiting}. We explore the normalization techniques in the proposed dense shortcut path mainly due to their lightweight property.

\subsection{Macro-architecture design}
On the other hand, network micro-architecture design aims to get backbone network structures that can be used to improve the performance across different tasks, for which the evaluation benchmark is the classification accuracy on ImageNet and CIFAR dataset, etc. Famous macro-architectures include AlexNet~\cite{krizhevsky2012imagenet}, VGGNet~\cite{simonyan2014very}, GoogleNet and its variants~\cite{szegedy2015going,szegedy2016rethinking}, ResNet~\cite{he2016deep}, and DenseNet~\cite{huang2017densely}. Being the pioneering networks in the deep learning field, AlexNet, VGGNet and GoogleNet are still widely used by many researchers to design a prototype network for their applications. However, there is a trend in the research community that ResNet and DenseNet have become more favourable choices due to their competitive performance and simple designs. ResNet has two famous variants: WideResnet~\cite{zagoruyko2016wide} and ResNext~\cite{xie2017aggregated}, which explore the dimensions of width and cardinality respectively. The original ResNet has demonstrated that identity shortcut can contribute to stabilizing the training of deep networks; however, the performance decreases when the network becomes extremely deep (for example, more than $200$ layers). Preactivation-ResNet~\cite{he2016identity} solved this problem by re-ordering activations in the ResNet module. The performance gain of preactivation-ResNet over original ResNet could only be observed in extremely deep networks~\cite{he2016identity}. Later, it has been found that the extreme depth is unnecessary since it provides worse performance compared with the performance by increasing the width of the network with a similar number of parameters~\cite{zagoruyko2016wide}. ResNext further explored the influence of cardinality, which is more effective than increasing either depth or width. Despite some small differences between ResNet and its two variants, \ie WideResnet and ResNext, one thing in common is that they all use identity shortcut. Instead, DenseNet effectively uses the concatenation technique without using the identity shortcut. Moreover, CondenseNet~\cite{huang2018condensenet}, a variant of DenseNet, exploits the power of dense connection in a more extreme way. One of its most significant differences from the DenseNet is that layers with different resolution feature-maps are also densely connected. Furthermore, Dual path network~\cite{chen2017dual} and mixed link network~\cite{wang2018mixed} integrate ResNet and DenseNet into one network by using both identity shortcut and dense concatenation. Our work differentiates from previous works in that our approach adopts neither identity shortcut nor dense concatenation. FractalNet~\cite{larsson2016fractalnet} also explored to train ultra-deep networks relying on neither identity shortcut nor dense concatenation; however, it provides less favourable performance. In the experiment section, we will show that our proposed approach outperforms both ResNet and DenseNet.

%-------------------------------------------------------------------------
\section{Proposed approach}

\subsection{Background: Dense connection exists in ResNet and DenseNet}

\begin{figure}[t]
\centering
\includegraphics[width=0.48\textwidth]{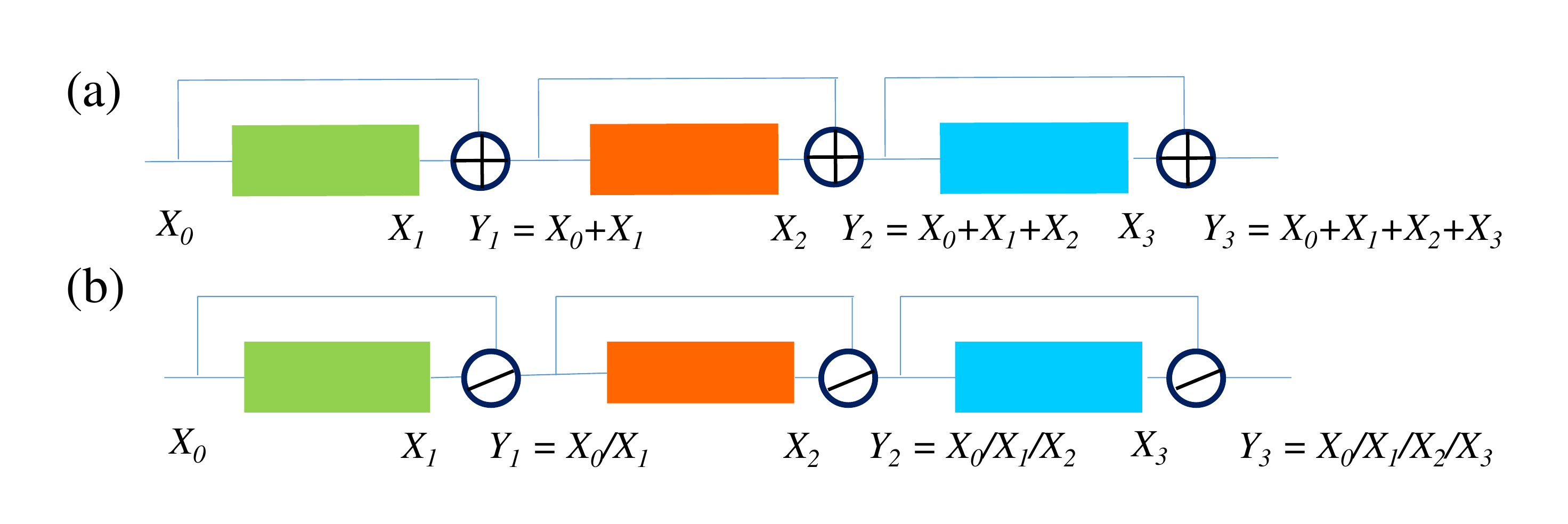}
%\vskip -0.2in
\caption{(a) ResNet and (b) DenseNet.}
\label{denseconnect}
\end{figure}

What is the difference between ResNet and DenseNet? As the name suggests, it seems that the difference lies in that ResNet only uses one preceding feature-map, while DenseNet uses features of all the preceding convolutional blocks. The shared philosophy that unifies ResNet and DenseNet is that they both connect to the feature-maps of all preceding convolutional blocks~\cite{huang2017densely}. A similar finding has been revealed in~\cite{chen2017dual,wang2018mixed}. That is to say, dense connection exists in both ResNet and DenseNet~\cite{chen2017dual,wang2018mixed}. For a typical convolution in DNNs, we formulate it as:
\begin{equation}
%\begin{gathered}
f_l = H_l*f_{l-1},\\
\label{eqn_basic}
%\end{gathered}
\end{equation}
where $f_l$ and $f_{l-1}$ indicate the current feature-map and previous feature-map respectively; ``$*$'' indicates convolution operation and $H_l$ indicates the convolution weight. For simplicity, we do not take the bias term in the convolution into account.
In VGG style network~\cite{simonyan2014very}, $f_{l-1}$ is only the preceding feature-map. In DenseNet, however, $f_{l-1}$ connects the feature-maps of all preceding convolutional blocks as illustrated in Figure\ \ref{denseconnect} (b):  
\begin{equation}
%\begin{gathered}
Y_l = X_0/X_1/.../X_l,\\
\label{eqn_concat}
%\end{gathered}
\end{equation}
where $X_i$ represents each of the preceding feature-maps and ``$/$'' represents the operation of concatenation. Replacing $f_{l-1}$ with above $Y_l$, we get
\begin{equation}
%\begin{gathered}
f_l = H_l*(X_0/X_1/.../X_l),\\
\label{eqn_conv_concat}
%\end{gathered}
\end{equation}
from which it is obvious that DenseNet uses the feature-maps of all preceding convolutional block outputs. In ResNet
\begin{equation}
%\begin{gathered}
Y_l = Y_{l-1} + X_l,\\
\label{eqn_resnet_basic}
%\end{gathered}
\end{equation}
and it may seem that $Y_l$ only reuses preceding feature-map $Y_{l-1}$. However, as shown in Figure\ \ref{denseconnect} (a) we can recursively extend this function and get
\begin{equation}
%\begin{gathered}
Y_l = X_0 + X_1 + ... + X_l,\\
\label{eqn_summation}
%\end{gathered}
\end{equation}
Likewise, we insert the above $Y_l$ to Eq.~\ref{eqn_basic}, for ResNet we get
\begin{equation}
%\begin{gathered}
f_l = H_l*(X_0 + X_1 +...+ X_l),\\
\label{eqn1_conv_summa}
%\end{gathered}
\end{equation}
from which it is clear that ResNet also connects to the feature-maps of all preceding convolutional blocks~\cite{huang2017densely,chen2017dual,wang2018mixed}. The difference between ResNet and DenseNet is that ResNet adopts summation to connect all preceding feature-maps while DenseNet concatenates all of them~\cite{wang2018mixed}.

\subsection{Unified perspective with dense summation}
\begin{figure}[t]
\centering
\includegraphics[width=0.48\textwidth]{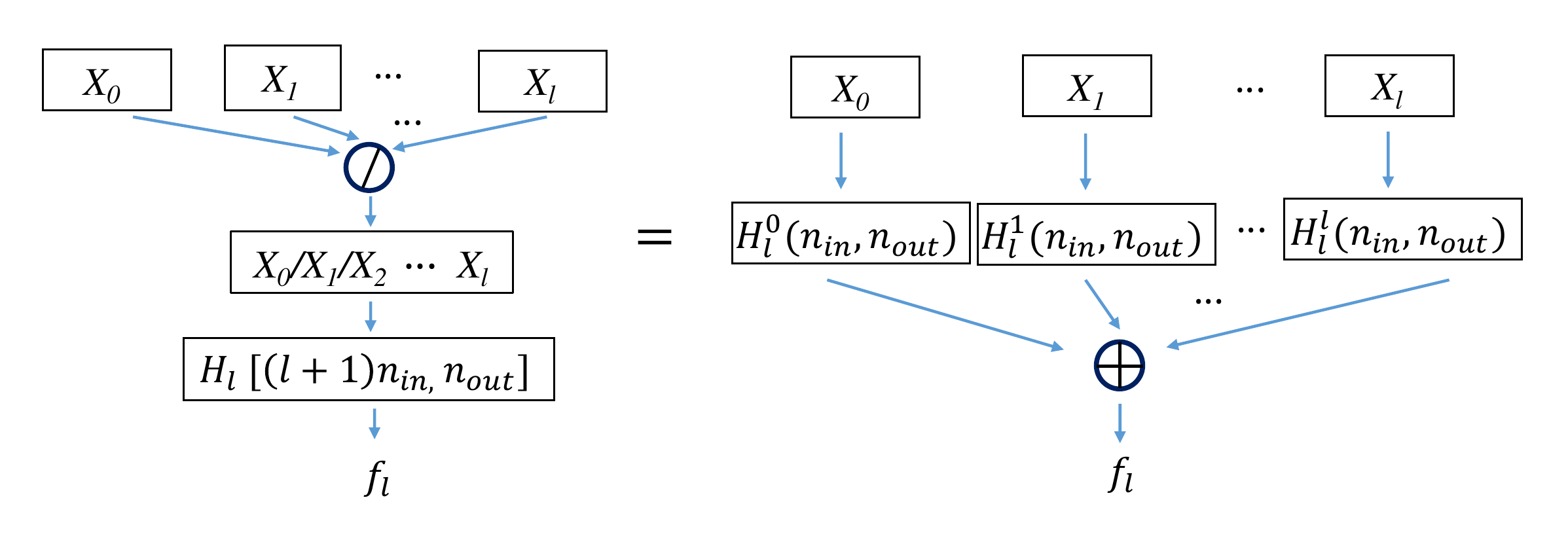}
%\vskip -0.2in
\caption{Equivalence of Dense concatenation before convolution and Dense summation after convolution.}
\label{conv_conc}
\end{figure}

As analyzed above, DenseNet is different from ResNet because they adopt different dense connection methods: summation vs concatenation. Here, we demonstrate that dense concatenation before convolution operation can be equivalent to dense summation after convolution, thus Eq.\ \ref{eqn_conv_concat} can be reformulated as:

\begin{equation}
%\begin{gathered}
\begin{split}
f_l = H_l*(X_0/X_1/.../X_l)\\
{} = (H_l^0/H_l^1/.../H_l^l)*(X_0/X_1/.../X_l)\\
{} = H_l^0*X_0 + H_l^1*X_1 + \dots + H_l^l*X_{l},
\label{eqn_dense_equi}
\end{split}
%\end{gathered}
\end{equation}
where $H_l = H_l^0/H_l^1/.../H_l^1$ simply dividing one convolution weight $H_l$ into multiple small convolution weights $H_l^i$, each having the same channel size as corresponding $X_i$. Note that $H_l$ indicates the first convolution in the convolutional block instead of the whole convolution block. Thus, $f_l$ is the feature-map after the first convolution in the convolutional block. Overall, the above equivalence is illustrated in Figure\ \ref{conv_conc}, where $n_{in}$ and $n_{out}$ are the number of input channels and output channels respectively. Similar observation has been revealed in~\cite{xie2017aggregated}. %With this equivalence, we can see that ResNet and DenseNet have even stronger connection with each other than the shared dense connection. 
For ResNet, we transform Eq.\ \ref{eqn1_conv_summa} into 
\begin{equation}
%\begin{gathered}
\begin{split}
f_l = H_l*(X_0 + X_1+...+ X_l)\\
{} = H_l*X_0 + H_l*X_1 + \dots + H_l*X_{l},
\label{eqn_resnet_equi}
\end{split}
%\end{gathered}
\end{equation}
We further summarize Eq.\ \ref{eqn_dense_equi} and Eq.\ \ref{eqn_resnet_equi} as follows.
\begin{equation}
%\begin{gathered}
\begin{split}
f_l = H_l^0*X_0 + H_l^1*X_1 + \dots + H_l^l*X_{l}\text{  for DenseNet,}\\
f_l = H_l*X_0 + H_l*X_1 + \dots + H_l*X_{l}\text{  for ResNet}.
\label{eqn_combined}
\end{split}
%\end{gathered}
\end{equation}

From Eq.\ \ref{eqn_combined} we observe that both ResNet and DenseNet have dense summation of $H_l(X_i)$. This interesting observation provides a more unified perspective to perceive ResNet and DenseNet in terms of their resemblance to each other. However, the main purpose of formulating this ``unified perspective” is to better understand their core differences for demonstrating their pros and cons. By comparing the two formulas for DenseNet and ResNet in Eq.\ \ref{eqn_combined}, we find that the core difference lies in that the convolution weight $H_l$ in ResNet is the same for every preceding layer output $X_i$ while $H_l^i$ is different for $X_i$ in DenseNet. This core difference results in other differences in practical use. The input channel $n_{in}*(l+1)$ in $H_l$ increases with the increase of $l$ and it is normally larger than that in $H_l$ of ResNet when $l$ becomes large. Due to the concatenation feature, $n_{out}$ in $H_l$ is normally very small but with more layers. Thus, DenseNet in practical use often requires more training time. Moreover, the concatenation nature resulting in large input channel $n_{in}*(l+1)$ also requires more GPU memory. However, the merit of DenseNet design is that it exhibits more flexibility of using previous feature-maps because each $H_l^i$ is different. 

It is worth mentioning that Eq.\ \ref{eqn_combined} does not reflect their practical implementation. $H_l*(X_0 / X_1 / \dots / X_{l})$ is much faster than $H_l^0*X_0 + H_l^1*X_1 + \dots + H_l^{l-1}*X_{l}$ on GPUs; and $H_l*(X_0 + X_1 + \dots + X_{l})$ is much more efficient than $H_l*X_0 + H_l*X_1 + \dots + H_l*X_{l}$. Our above analysis only demonstrates the theoretical connection between ResNet and DenseNet.

\subsection{Dense shortcut and DSNet}

\begin{figure}[t]
\centering
\includegraphics[width=0.48\textwidth]{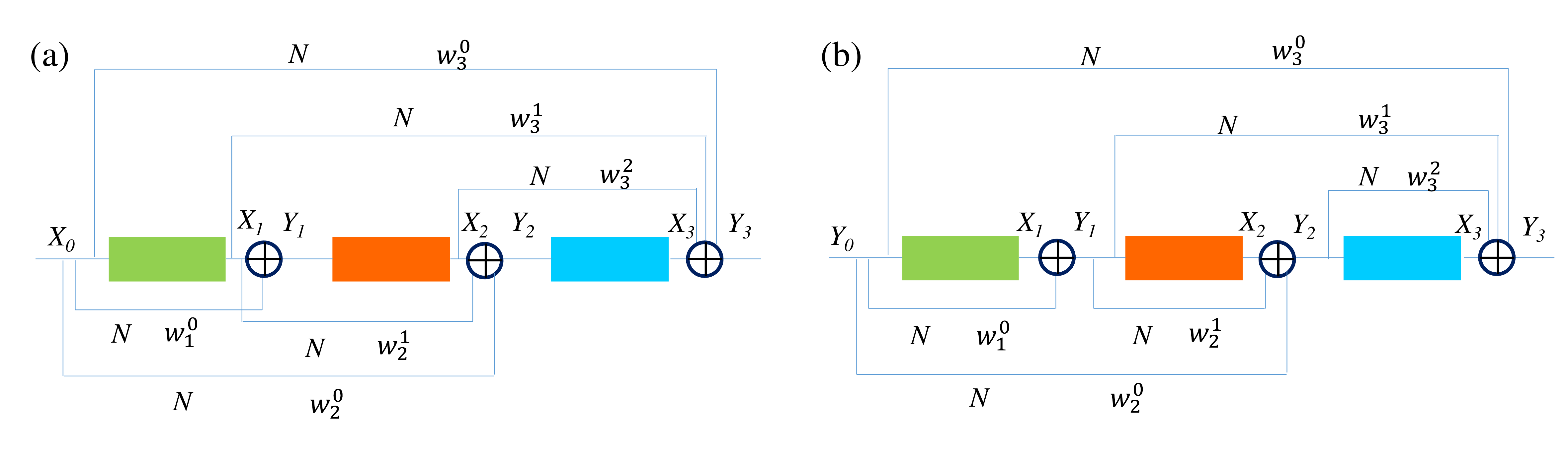}
%\vskip -0.2in
\caption{Proposed DSNet adopting dense (weighted normalized) shortcut. (a) Densely connected to $X_l$, (b) Densely connected to $Y_l$.}
\label{denseshortcut}
\end{figure}

With the above unified perspective, the core difference between ResNet and DenseNet is revealed as whether the convolution parameters $H_l$ are shared for each preceding output. It then results in superior performance of DenseNet with the disadvantage of requiring more GPU resources. The difference originates from the adoption of different dense connection techniques, identity shortcut and dense concatenation. In this paper we propose one alternative dense connection that is motivated to alleviate their drawbacks. It introduces flexibility of using preceding feature-maps while still using the same $H_l$ for each preceding feature-map. Benchmarking ResNet formula in Eq.\ \ref{eqn_combined}, we propose
\begin{equation}
%\begin{gathered}
\begin{split}
f_l = H_l*DS^{0}_{l}(X_0) + H_l*DS^{1}_{l}(X_1) + \dots \\ 
+ H_l*DS^{l-1}_{l}(X_{l-1}) + H_l*X_{l} .
\label{eqn_from_x}
\end{split}
%\end{gathered}
\end{equation}
which is equivalent to
\begin{equation}
%\begin{gathered}
\begin{split}
f_l = H_l*(DS^{0}_{l}(X_0) + DS^{1}_{l}(X_1) + \dots \\ 
+ DS^{l-1}_{l}(X_{l-1}) + X_{l}),
\label{eqn_from_x_new}
\end{split}
%\end{gathered}
\end{equation}
where ``$DS()$'' indicates dense shortcut. It refers to dense weighted normalized shortcut consisting of normalization and channel-wise weight. Specifically, $DS^{i}_{l}(X_i) = W^{i}_{l} \times N(X_i)$, where $W^{i}_{l}$ represents channel-wise weight and and $N$ indicates normalization operation. In this work, the term ``dense shortcut" is equivalent to ``dense weighted normalized shortcut" unless otherwise specified. Eq.\ \ref{eqn_from_x_new} is illustrated in Figure\ \ref{denseshortcut} (a); however, we conjecture that the feature-map contains more useful feature in the aggregation output $Y_l$ than its corresponding single convolutional block output $X_l$, thus by replacing $X_l$ with $Y_l$ we propose another variant:
\begin{equation}
%\begin{gathered}
\begin{split}
f_l = H_l*(DS^{0}_{l}(Y_0) + DS^{1}_{l}(Y_1) + \dots \\ 
+ DS^{l-1}_{l}(Y_{l-1}) + X_{l}),
\label{eqn5}
\end{split}
%\end{gathered}
\end{equation}
which is illustrated in Figure\ \ref{denseshortcut} (b). The conjecture that the feature of aggregation output $Y_l$ is more meaningful than $X_l$ is supported by the experimental results (see Table~\ref{table1}). Therefore we mainly adopt variant (b) in Figure\ \ref{denseshortcut} in this study. We term the proposed network adopting DS shortcut DSNet. DSNet adopts the same network backbone (convolutional block itself and block design) as ResNet~\cite{he2016deep}. The ResNet backbone is tailored for the identity shortcut, not for our proposed shortcut. It is conceivable that redesigning the backbone structure might further improve the performance of our DSNet, but this is beyond the scope of this work. The only difference between our DSNet and ResNet~\cite{he2016deep} is to replace identity shortcut with the proposed dense shortcut (ie., the dense weighted normalized shortcut).

For introducing dense shortcuts in ResNet, one naive approach is to just densely connect all preceding feature-maps by replacing single identity shortcut in Eq.\ \ref{eqn_resnet_basic} with dense identity shortcut, and we get
\begin{equation}
%\begin{gathered}
Y_l = Y_{l-1} + Y_{l-2}+...+ Y_{0} + X_l,\\
\label{eqn2}
%\end{gathered}
\end{equation}
which can be recursively extended as:
\begin{equation}
%\begin{gathered}
Y_l = X_l + X_{l-1} + 2X_{l-2} +...+(l-1)X_{1} + lX_{0}.\\
\label{eqn2_new}
%\end{gathered}
\end{equation}

Comparing it with Eq.\ \ref{eqn_summation}, we find that dense identity shortcut is equivalent to add extra constant at the end of each convolutional block. Such design denoted ResNet50-dense in Table~\ref{table1} does not achieve better performance than original ResNet50. This demonstrates the failure of naive dense identity shortcut. Next, we will illustrate our motivation for the design of our DS shortcut.

The motivation of using normalization is to normalize all the preceding features into a similar scale to avoid any preceding feature to dominate the whole summation and facilitate the training. Note that no affine transformation is applied in the normalization process. The weighted summation is to provide the network freedom to assign proper weight to each normalized feature-map depending on its significance. It is cumbersome to manually decide the weights for each one, thus these weights are set to learnable parameters. Moreover, we empirically find that inserting weighted normalized shortcut within the convolutional block on the 3$\times$3 convolution can also contribute to the performance improvement, and since it adds almost little computation burden, it is worth including it. We term it DS2Net.

\subsection{Ablation study}

We adopt the widely used ResNet50 backbone to do ablation studies.
The test is conducted on CIFAR100 and the results are available in Table~\ref{table1}. Note that the width of the network is set to 0.25 times the width of ResNet in~\cite{he2016deep} to save computation resources. We have two major observations from Table~\ref{table1}. First, normalization techniques are important to improve performance. Specifically, group normalization (GN) outperforms batch normalization (BN) by a visible margin. Instance normalization (IN) and layer normalization (LN) are two special cases of GN. LN performs slightly inferior to GN, and IN performs the worst. Second, the weighted parameters are also critical to improve performance. In particular, we empirically find that adopting channel-wise weight is crucial for the performance gain which indicates that different channels should have different weights. Overall, these two observations support the intuitions of adopting weighted normalized shortcuts. Other observations include the superiority of DS2Net to DSNet and the inferiority of DSNet-a to DSNet.  
\begin{table}[t]
\centering
%\vskip -0.2in
%\begin{minipage}[t]{0.52\textwidth}
\centering
%
%\begin{table}
%\begin{center}
\begin{small}
\caption{Classification error (\%) on CIFAR-100 validation dataset for ablation study, ``-a" indicates structure (a) in Figure\ \ref{denseshortcut}, others by default adopts structure (b) in Figure\ \ref{denseshortcut}.}
\label{table1}
\scalebox{0.88}{
\begin{tabular}{|l|c|c|c|c|}
\hline
Architecture & Top-1 (\%) \\
\hline\hline
ResNet50 (0.25$\times$) \cite{he2016deep}  & 26.13 \\ 
ResNet50-dense (0.25$\times$)  & 26.33 \\ 
DSNet50-a (GN + weight) (0.25$\times$)   & 25.23 \\ 
DSNet50 (GN + weight) (0.25$\times$)   & 24.12 \\ 
DS2Net50 (GN + weight) (0.25$\times$)   & 23.72  \\ 
DSNet50 (LN + weight) (0.2$\times$)  & 24.43 \\ 
DSNet50 (BN + weight) (0.25$\times$)  & 24.90 \\ 
DSNet50 (IN + weight) (0.25$\times$)  & 28.11 \\ 
DSNet50 (None + weight) (0.25$\times$) & 26.26 \\ 
DSNet50 (GN + no weight) (0.25$\times$)     & 25.85 \\  

\hline
\end{tabular}
}

\end{small}
%\end{center}
\vskip 0.05in

\end{table}

\begin{table}[t]
\centering
%\vskip -0.2in
%\begin{minipage}[t]{0.52\textwidth}
\centering
%
%\begin{table}
%\begin{center}
\begin{small}

\caption{Classification error (\%) on CIFAR-100 and CIFAR-10 validation dataset for different widths and depths.}
\label{table2}
\scalebox{0.88}{
\begin{tabular}{|l|c|c|c|c|}
\hline
Architecture & CIFAR-100 & CIFAR-10 \\
\hline\hline
ResNet50 (1$\times$) \cite{he2016deep}  & 21.43 & 4.82\\ 
DSNet50 (1$\times$) & 19.95 & 4.54\\
DS2Net50 (1$\times$)   & 19.00 & 4.33 \\ 
ResNet50 (0.25$\times$) & 26.13 & 6.65\\
DSNet50 (0.25$\times$) & 24.12 & 5.95\\
DS2Net50 (0.25$\times$)  & 23.72 &5.77\\ 
ResNet101 (0.25$\times$) & 25.00 & 6.00\\
DSNet101 (0.25$\times$) & 23.94 & 5.82\\
DS2Net101 (0.25$\times$)  & 23.61 & 5.68\\ 

\hline
\end{tabular}
}
\end{small}
%\end{center}
\vskip 0.05in

\end{table}

\section{Experimental Results and Analysis}
In this section, we conduct experiments across a range of datasets to evaluate the proposed dense shortcut.

\subsection{CIFAR experiments}
We first evaluate it on CIFAR datasets. Both CIFAR-100 and CIFAR-10 contain 60,000 colour images with the resolution of 32 $\times$ 32. Totally 50,000 images are used as the training dataset and the remaining 10,000 images as the validation dataset. For the data augmentation, we adopt widely adopted cropping with 4-pixel padding and horizontal flipping. The preprocessing is done with the normalization through the training dataset mean and standard deviations values. For all CIFAR experiments, by default, we set the weight decay to 0.0005 and train for 64k iterations and the initial learning rate is set to 0.1  which is then divided by 10 at 32k and 48k iterations respectively similar to~\cite{he2016deep}. Since we adopt the same backbone as ResNet and replace the identity shortcut with our dense shortcut, we mainly compare our results with the original ResNets~\cite{he2016deep}. 

The results in Table~\ref{table2} show that our proposed approach outperforms ResNet by a significant margin. On CIFAR-100, DSNet50 ($\times$0.25) outperforms ResNet50 ($\times$0.25) by a margin of 2\%. DS2Net further improves the performance of DSNet by a visible margin. The same trend can be observed for both wider ($1\times$) and deeper networks (depth 101). CIFAR-10 mirrors the story and for simplicity in the remainder of the paper, we only report the result for CIFAR-100. To further evaluate the robustness of the proposed DSNet to different depths and widths, we conduct extra experiments and the results are available in Table~\ref{table3} and Table~\ref{table4} respectively. 

\begin{table}[!htbp]
\centering
%\vskip -0.2in
%\begin{minipage}[t]{0.52\textwidth}
\centering
%
%\begin{table}
%\begin{center}
\begin{small}
\caption{Classification error (\%) on CIFAR-100 validation dataset for different depths with the width 0.25$\times$.}
\label{table3}
\begin{tabular}{|l|c|c|c|c|}
\hline
Depth & block design & ResNet & DSNet & DS2Net \\
\hline\hline
26 & [2, 2, 2, 2]& 27.74 & 26.80 & 26.30 \\ 
38 & [3, 3, 3, 3]& 27.26 & 25.59 & 25.30 \\ 
50 & [3, 4, 6, 3]& 26.13 & 24.12 & 23.72 \\ 
%62 & [3, 4, 10, 3]& 25.71 & ? & ? \\ 
77 & [3, 4, 15, 3]& 25.32 & 23.95 & 23.55 \\ 
101 & [3, 4, 23, 3]& 25.01 & 23.63 & 23.40 \\ 

\hline
\end{tabular}
\end{small}
%\end{center}
% \vskip 0.05in
\end{table}

\begin{table}[!htbp]
\centering
%\vskip -0.2in
%\begin{minipage}[t]{0.52\textwidth}
\centering
%
%\begin{table}
%\begin{center}
\begin{small}
\caption{Classification error (\%) on CIFAR-100 validation dataset for different widths with ResNet as the backbone.}
\label{table4}
\scalebox{0.88}{
\begin{tabular}{|l|c|c|c|c|}
\hline
Width & ResNet50 & DSNet50 & DS2Net50 \\
\hline\hline
0.25 & 26.13 & 24.12 & 23.72 \\ 
0.25(WRN) & 23.51 & 22.00 & 21.58 \\ 
0.5 & 23.38 & 21.59 & 21.02 \\ 
1.0 & 21.43 & 19.95 & 19.00 \\ 

\hline
\end{tabular}
}
\end{small}
%\end{center}
%\vskip 0.05in

\end{table}

For the depth exploration, we set the width to $0.25\times$ and for the width exploration, we set the depth to 50 layers. We observe that DSNet consistently outperforms ResNet over a wide range of depth and widths. Since it has been shown by~\cite{zagoruyko2016wide} that increasing the width of the network is more effective than increasing the depth to improve the performance, in the following exploration, we always choose the depth to be 50 layers. Furthermore, we evaluate the proposed dense shortcut on one famous variant of Resnet: ResNext. The results are available in Table~\ref{table5}. The results show that a similar trend has been observed as the original ResNet. Note that ResNext has a similar number of parameters as ResNet, while wide ResNet has almost three times more parameters and GFLOPS. Since the wide ResNets essentially adopts the same structure as ResNet and the small difference is only doubling the Conv 3$\times$3 feature-maps, to avoid redundancy we only report one case in Table~\ref{table4} and do not perform more experiments on this structure. Even though ResNeXt is a well-optimized structure, our proposed approach can further improve its performance with a significant margin. Interestingly, DS2NeXt (0.5$\times$) can even outperform ResNeXt (1$\times$) with a margin of 0.66\%. This is quite surprising because ResNeXt (1$\times$) has around four times more parameters and GFLOPS than DS2NeXt (0.5$\times$). Note that the number of parameters and GFLOPS increase linearly with the increase of depth but quadratically with the increase of width. We further compare our performance on CIFAR-100 with the performance reported in previous works. WRN-28-10 does not adopt the bottleneck structure, thus DS2WRN-28-10 is not applicable (the shortcut within the Conv block can only be inserted into the Conv 3$\times$3 in the bottleneck for the dimension to match). From Table~\ref{table6} we observe that the networks that do not use dense connection perform much worse than those adopting dense connection. FractalNet adopting neither identity shortcut nor dense concatenation but with careful engineering design performs much better than other networks without dense connection; however, the performance is still not comparable to ResNet variants or DenseNets. Similar to FractalNet our proposed DSNet adopts neither identity shortcut nor dense concatenation, but it outperforms both ResNets (including WRN and ResNeXt) and DenseNets. The results show that dense weight normalized shortcuts constitute a competitive dense connection technique.

\begin{table}[t]
\centering
%\vskip -0.2in
%\begin{minipage}[t]{0.52\textwidth}
\centering
%
%\begin{table}
%\begin{center}
\begin{small}
\caption{Classification error (\%) on CIFAR-100 validation dataset for different widths with ResNeXt as the backbone.}
\label{table5}
\scalebox{0.88}{
\begin{tabular}{|l|c|c|c|c|}
\hline
Width & ResNeXt50 & DSNeXt50 & DS2NeXt50 \\
\hline\hline
0.25 & 23.86 & 22.98 & 22.58 \\ 
0.5 & 21.16 & 19.95 & 19.51 \\ 
1.0 & 20.17 & 18.58 & 18.24 \\ 

\hline
\end{tabular}
}
\end{small}
%\end{center}
% \vskip 0.05in

\end{table}

\begin{table}[t]
\centering
%\vskip -0.2in
%\begin{minipage}[t]{0.52\textwidth}
\centering
%
%\begin{table}
%\begin{center}
\begin{small}
\caption{Classification error (\%) on CIFAR-100 validation dataset.}
\label{table6}
\scalebox{0.77}{
\begin{tabular}{|l|c|c|c|c|}
\hline
Architecture & params & Top-1 (\%) \\
\hline\hline
ALL-CNN~\cite{springenberg2014striving}  & - & 33.71\\ 
Deeply supervised Net~\cite{lee2015deeply}& - & 34.57\\
HighWay Network~\cite{srivastava2015highway}   & - & 32.39 \\ 
FractalNet~\cite{larsson2016fractalnet} & 38.6M & 23.30 \\
with dropout~\cite{larsson2016fractalnet} & 38.6.M & 23.73 \\
ResNet~\cite{he2016deep} & 1.7M & 27.22\\
ResNet with stochastic depth~\cite{huang2016deep} & 1.7M & 24.58\\
Preacitvation ResNet~\cite{he2016identity} & 10.2M & 22.71 \\
DenseNet($k$ = 24)~\cite{huang2017densely} & 27.2M & 19.25\\
DenseNet-BC ($k$ = 24)~\cite{huang2017densely} & 15.3M & 17.60\\ 
DenseNet-BC ($k$ = 40)~\cite{huang2017densely} & 25.6M & 17.18\\
\hline
WRN-28-10~\cite{zagoruyko2016wide} & 36.5M & 19.25 \\
with dropout~\cite{zagoruyko2016wide} & 36.5M & 18.85\\
ResNeXt-29, 8$\times$64$d$~\cite{xie2017aggregated} & 34.4M & 17.77\\
ResNeXt-29, 16$\times$64$d$~\cite{xie2017aggregated} & 68.1M & 17.31\\
\hline
DSWRN-28-10 & 36.5M & 18.33\\
DSNeXt-29, 6$\times$64$d$ & 34.4M & 16.85\\
DS2NeXt-29, 6$\times$64$d$ & 34.4M & 16.39\\
\hline
\end{tabular}
}
\end{small}
%\end{center}
\vskip 0.05in

\end{table}

\subsection{ImageNet experiments}
ImageNet is the benchmark dataset for classification tasks to evaluate and compare different approaches. Our implementation details follow ResNet~\cite{he2016deep}. Specifically, we adopt the commonly used random $224\times224$ cropping with scale and aspect ratio augmentation for training and adopt SGD as the optimizer. For typical training on ImageNet, 8 GPUs are used and the batch size is set to 256~\cite{he2016deep}. We used 4 GPUs to train the proposed DSNet and the batch size is set to 128 (32 per GPU). Accordingly, taking linear scaling rule into account, we set the initial learning rate to 0.05 instead of the commonly used 0.1. We train the network for 100 epochs and the learning rate is divided by 10 after every 30 epochs. We report the single-crop classification errors on the validation dataset with the input image size of $224\times224$. Both top-1 and top-5 errors are reported and the results are available in Table \ref{table7}. 

\begin{table}[!htbp]
\centering
%\vskip -0.2in
%\begin{minipage}[t]{0.52\textwidth}
\centering
%
%\begin{table}
%\begin{center}
\begin{small}

\caption{Classification top-1 error (\%) on ImageNet validation dataset.}
\label{table7}
\scalebox{0.77}{
\begin{tabular}{|l|c|c|c|c|}
\hline
Architecture & Params & Top-1 (\%) & Top-5 (\%)\\
\hline\hline
ResNet50 \cite{he2016deep}   &25.6M & 24.01 & 7.02\\ 
ResNet101 \cite{he2016deep}   &44.6M & 22.44 & 6.21\\ 
ResNet152 \cite{he2016deep}   &60.2M & 22.16 & 6.16\\ 
WRN-50-2-bottleneck \cite{zagoruyko2016wide}  &69.8M & 21.9 & 6.03\\
ResNeXt-50, 32$\times$4$d$ \cite{zagoruyko2016wide}  &25M & 22.2 & -\\
SE-ResNet50 \cite{hu2018squeeze}  &28.1M & 23.29 & 6.62\\
CBAM-ResNet50 \cite{woo2018cbam}  &28.1M & 22.66 & 6.31\\
\emph{b}-RGSNet50 \cite{zhang2019revisiting} & 25.6M & 22.68 & 6.42\\
Res-RGSNet50 \cite{zhang2019revisiting} & 25.6M &22.21 & 5.99\\
DenseNet201 \cite{huang2017densely}  &20M & 22.58 & 6.34\\
DenseNet264 \cite{huang2017densely}  &33.3M & 22.15 & 6.12\\
Res2Net \cite{gao2019res2net}  &33.3M & 22.01 & 6.15\\
\hline
DSNet50    &25.6M & 22.49 & 6.29 \\ 
DS2Net50   &25.6M & 22.03 & 5.93 \\ 
DS2Res2Net50   &25.6M & 21.61 & 5.83 \\ 

\hline
\end{tabular}
}
\end{small}
%\end{center}
\vskip 0.05in

\end{table}

\begin{figure}[!htbp]
    \centering
    \includegraphics[width=0.47\textwidth]{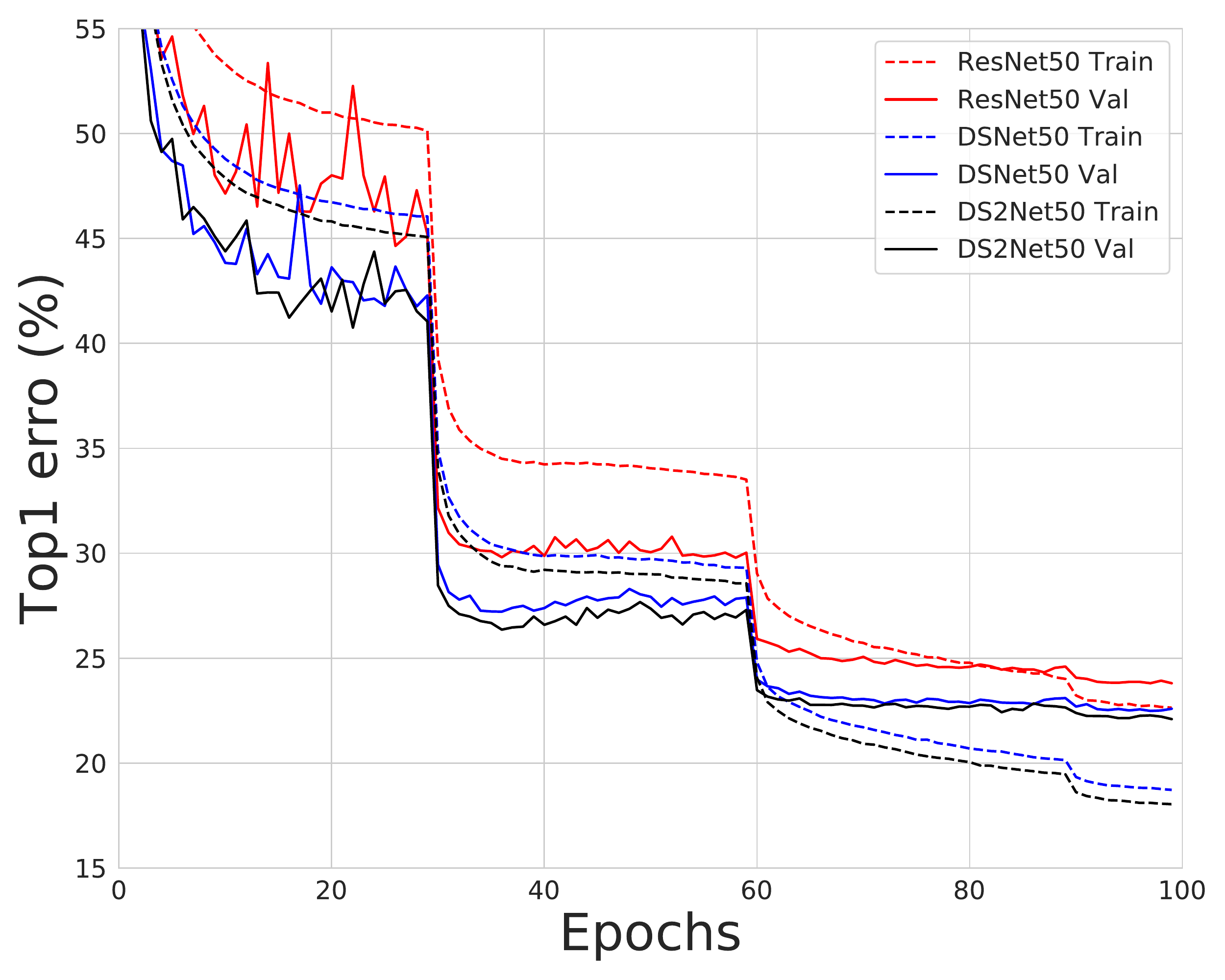}
    \caption{Training curves on ImageNet.}
    \label{traincurve}
\vspace{-2mm}
\end{figure}

We note that the proposed DS2Net50 can achieve significantly better performance than the original ResNet50. Somewhat surprisingly, DS2Net50 can outperform SE-Net\cite{hu2018squeeze} (which won the first place in ILSVRC 2017 challenge) as well as CBAM with improved attention module by a relatively large margin. DS2Net50 can achieve equivalent (if not better) performance with that of much deeper ResNet152. Compared with WRN-50-2-bottleneck, DS2Net achieves slightly worse result with the top-1 error metric and slightly better performance with the top-5 error metric. Note that WRN-50-2 has almost three times more parameters and GFLOPS than DS2Net. DS2Net-50 achieves slightly better performance than ResNeXt-50, 32$\times$4$d$ and Res-RGSNet50. It is claimed by \cite{huang2017densely} that DSNet201 can achieve equivalent performance with ResNet101 which has twice more parameters and computation GFLOPS. Even though DS2Net with a smaller number of parameters or GFLOPS outperforms DenseNet264, we do not aim to argue it as the main merit of DS2Net over DenseNet. Instead, the main advantage of DS2Net over DenseNet is that it is more time-efficient in practice with GPU implementation. In short, DSNet shows equivalent or better performance with DenseNet but it avoids the drawback of DenseNet. Our proposed DS2Net-50 also achieves comparable performance as the recent Rese2Net-50~\cite{gao2019res2net}. By applying the dense shortcuts to Res2Net, we further improve the performance by 0.40\% margin, which is not trivial considering Res2Net is already a very well designed architecture. Compared with ResNet50, DS2Net50 improves the performance of ResNet50 by a large margin. The number of the added weighted parameters is less than 0.15\% of the original number of parameters and only a small computation overhead is added because it still adopts the same backbone structure as the original ResNet and only light operations are needed for the added shortcut. \\
The training curve is shown in Figure\ \ref{traincurve}. We observe that the proposed dense weighted normalized shortcut is also beneficial to speed up the convergence. It is also important to note that the training error of DS2Net is much smaller.

\begin{figure*}[!htbp]
    \centering
    \includegraphics[width=0.80\linewidth]{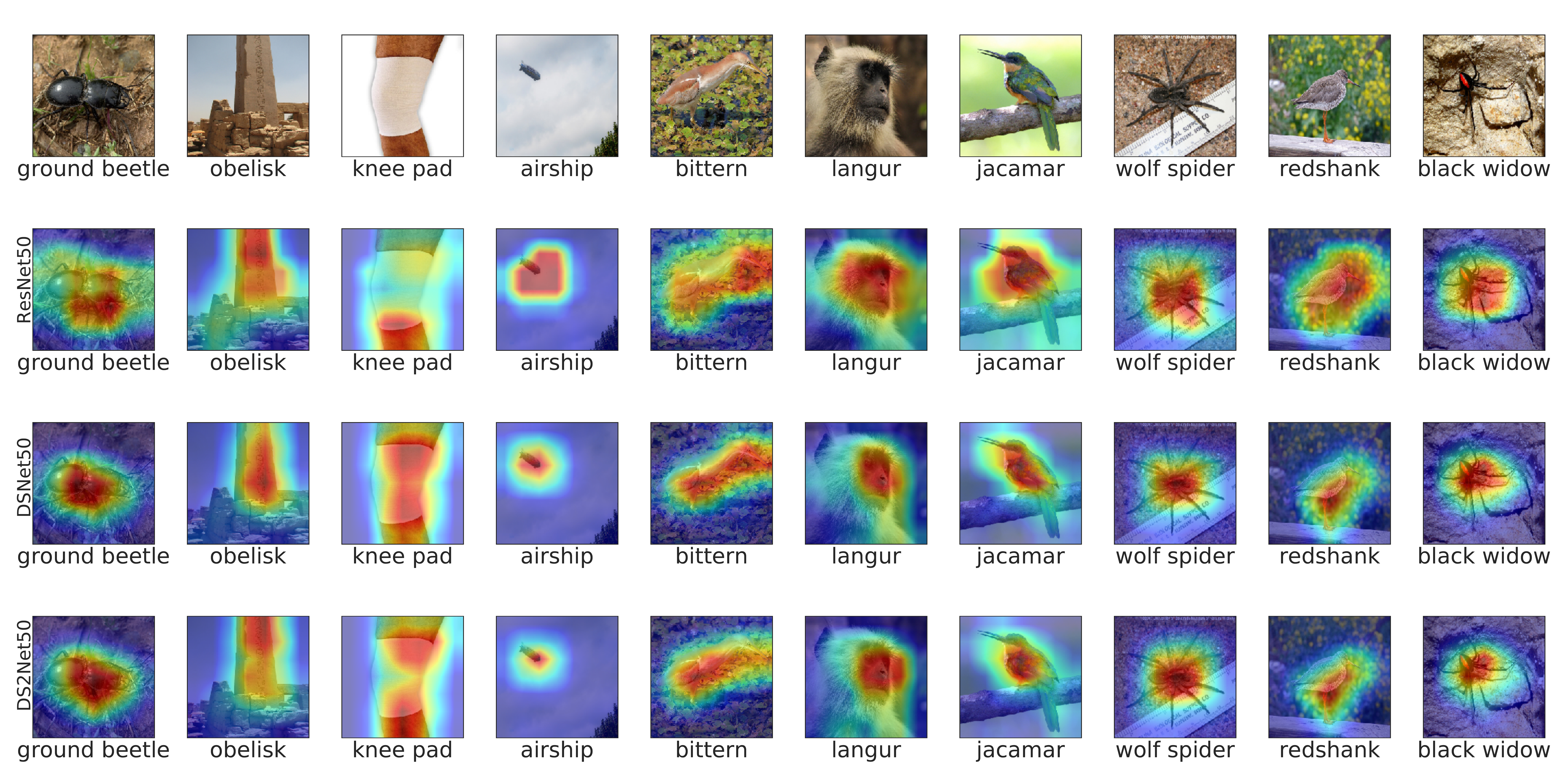}
    \caption{Grad-CAM~\cite{selvaraju2017grad} visualization results of ResNet50 (second row), DSNet50 (third row) and DS2Net50 (llast row).}
    \label{fig:gradcam}
\vspace{-2mm}
\end{figure*}

\subsection{COCO detection dataset experiments}

To evaluate the dataset generalization capability of the proposed DSNet, we further evaluate it on MS COCO 2014 detection dataset~\cite{lin2014microsoft}. Faster-RCNN~\cite{ren2015faster} is chosen as the detection method. The network is pre-trained on ImageNet and then finetuned on the COCO dataset with 5 epochs for fast performance validation. The results are available in Table \ref{table8}. On COCO detection dataset, our proposed DSNet achieves better performance than ResNet50.

\begin{table}[!htbp]
\centering
%\vskip -0.2in
%\begin{minipage}[t]{0.52\textwidth}
\centering
%
%\begin{table}
%\begin{center}
\begin{small}
\caption{mAP (\%) on MS COCO validation dataset.}
\label{table8}
\scalebox{0.77}{
\begin{tabular}{|l|c|c|c|c|}
\hline
Backbone & mAP\@.5 & mAP\@.75 & mAP\@ [.5, .95]\\
\hline\hline
ResNet50     & 51.3 & 33.6 & 31.4\\
DSNet50      & 54.2 & 36.0 & 33.7\\
DS2Net50       & 54.3 & 36.2 & 33.7\\
\hline
\end{tabular}
}

\end{small}
%\end{center}
\vskip 0.05in

\end{table}

\subsection{Visualization with Grad-CAM}
We apply the widely used Grad-CAM~\cite{selvaraju2017grad} to ResNet50~\cite{he2016deep} and our proposed DSNet/DS2Net, on the images from the ImageNet validation set. The visualization results are available in Figure~\ref{fig:gradcam}. Grad-CAM calculates the gradients concerning a certain class, thus Grad-CAM result shows the attended regions in the image. We observe that the DSNet attends more on the objects and have a sharper focus (i.e. attention) than ResNet50. No obvious difference is observed between DSNet50 and DS2Net50. More qualitative results are available in the supplementary material.

\subsection{Implementation design and memory/speed test}

The straightforward implementation of dense normalized shortcut is to do normalization with affine transformation in every shortcut. This will cause unnecessary computation and slightly more parameter overhead burden. First, the normalization process is shared for a certain preceding layer feature. Second, the affine transformation by default has both scale and bias, while the summation of dense bias is mathematically redundant. In our implementation, for every aggregation output $Y_l$, we only perform the normalization process once which is then shared by all the dense shortcuts linked to it. Thus, for the dense shortcut path, the operation is only to multiply the shared normalized feature by the corresponding weight, which is very light. This implementation choice does not influence the performance but requires less computation time.

\begin{table}[!htbp]
\centering
%\vskip -0.2in
%\begin{minipage}[t]{0.52\textwidth}
\centering
%
%\begin{table}
%\begin{center}
\begin{small}

\caption{GPU memory and training time on ImageNet; memory indicates that per GPU and time indicates that per iteration.}
\label{table9}
\scalebox{0.77}{
\begin{tabular}{|l|c|c|c|c|}
\hline
Architecture & Top-1 (\%) & memory (MB) & time (s)\\
\hline\hline
ResNet50 \cite{he2016deep}   &24.01 & 3929 & 0.31\\ 
ResNet152 \cite{he2016deep}   &22.16 & 7095 & 0.63\\ 
DenseNet264 \cite{huang2017densely}  & 22.15 & 9981 & 0.60\\
\hline
DSNet50    &22.49 & 4777 & 0.37 \\ 
DS2Net50   &22.03 & 5133 & 0.39 \\ 

\hline
\end{tabular}
}
\end{small}
%\end{center}
\vskip 0.05in
\end{table}

With the above implementation, we measure the ImageNet training memory and speed on the same machine (equipped with four 1080Ti GPUs) with the hyperparameters specified above. We report consumed memory per GPU and computation time per iteration in Table\ \ref{table9}. DS2Net (with a smaller number of parameters as shown in Table\ \ref{table7}) performs relatively better than ResNet152 and Dense264 with the advantage for both memory and speed. Compared with ResNet50, the increase of memory and computation for DSNet/DS2Net is marginal. 

\section{Conclusions}
We provide a unified perspective of dense summation to facilitate the understanding of the core difference between ResNet and DenseNet. We demonstrate that the core difference lies in whether the convolution parameters are shared for the preceding feature-maps. We proposed a dense weighted normalized shortcut as an alternative dense connection method, which outperforms the two existing dense connection techniques: identity shortcut in ResNet and dense concatenation in DenseNet. We found that Dense summation from the aggregation output provides superior performance to that from the convolutional block output. In short, the dense shortcut addresses the problem of representational capacity decrease in ResNet while avoiding the drawback of requiring more GPU resources in DenseNet. The proposed DSNet has been evaluated on multiple benchmark datasets to show superior performance than its counterpart ResNet. For example, on ImageNet, DenseNet50 can achieve better performance than the much deeper ResNet152. On ImageNet with few parameters and computation, it also achieves comparable performance as DenseNet. Moreover, it also achieves comparable performance as the recent Res2Net which can be further boosted by our dense shortcuts. %In this work, we only introduce dense shortcut to ResNet but not to DenseNet, because adopting a backbone of DenseNet will require more computation resource and offset the benefit the dense shortcuts bring. 
Compared with other ``free" performance boost module such as SE and CBAM, our dense shortcut also achieves more superior performance. The Grad-CAM result shows that DSNet, in general, focuses better on the object in the image than its counterpart ResNet.

{\small
\bibliographystyle{ieee_fullname}
\bibliography{egbib}
}

\end{document}